# IS (Iris Security)

**Gerardo Iovane, Francesco Saverio Tortoriello**
**Researchers**
**Dipartimento di Ingegneria Informatica e Matematica Applicata – Università degli Studi di Salerno**

Research, Imaging Equipment, University/Education

## Products used

- LABVIEW Prof Dev Sys 6.1
- LABVIEW Real Time Module 6.1
- IMAQ Vision 6.01
- SQL TOOLKIT 2.0
- INTERNET TOOLKIT 5.0
- SIGNAL PROCESSING TOOLSET
- SPC Tools

**The challenge:** The development of system, based on iris analysis, for people identifying and recognizing where it is required an high level of security.
.

**The solution:** A real time vision system is projected and implemented with innovative algorithms to execute the recognition in multidimensional spaces and transformed ones respect to the space of the native image.

## Abstract
In the paper will be presented a safety system based on iridology. The results suggest a new scenario where the security problem in supervised and unsupervised areas can be treat with the present system and the iris image recognition.

## Introduction
One of the most relevant question in the security world is how identify and recognize people not only in some protected areas with a supervisor, but also in public area with high people density like as in airports or banks. Some biometric experiments are made and others are examining such as the fingerprint. The most important limits of these system are: 1) the more trivial systems can confuse people with a probability of about 50%, whereas the identification of the same person in the more sophisticate systems could fail with a probability of the 10%. Nowadays for this reason the automatic identifying systems are not so used. Here I present a system based on the iridology, which has a scaleable accuracy respect to different CCDs and so respect to the request and budget. In particular the present system reduces the failure of the system at the $10^{-7}$ % respect the person mistake and $10^{-6}$ % respect to the not recognize of the same person. Thanks to this results the present iris vision system could be used not only in supervised environment but also for general purposes, where at the moment smart cards are used.

## The IRIS System

The hardware consists of a standard pc with the Microsoft Window as OS, the NI 1411 acquisition board, and an analogic colour single chip CCD camera. The system architecture is structured as shown in Fig.1. The software is realized with Labview and Labview RT. The image analysis is realized with IMAQ and the data analysis use the SIGNAL PROCESSING TOOLSET by NI. A private library was developed for the recognition based on wavelet analysis, neural network and genetic algorithms.

**i)** The Image acquisition unit is responsible for the image acquisition and pre-reduction (geometrical calibration, photometric alignment).

**ii)** The supervisor can control the work of the automatic system by the console, which is also employed for the registration of new people/users (see Fig.2).

**iii)** The raw images are temporary stored in the DataBase (DB) Unit (see Figs.3, 4). It uses SQL language, and is linked to the system by SQL TOOLKIT by NI.

**iv)** Each image is given to the Process and Analysis Unit (P&A) for the recognition. This unit is formed by four subunits: 1) the first subunit splits the colored image in the RGB frames and tests the morphology of the eyes in the real space; 2) the second subunit transforms the image in a 3-d image, where the third dimension corresponds to a different weight of the eye respect to a fixed coordinate frame and respect to some characteristic parameters in iridology (see Fig.5); 3) the third subunit transforms the image in the frequencies domain where a wavelets analysis is carried out (here the best coefficients are fixed thanks to a neural network based on a multilayer perceptron); 4) the last subunit transforms the image in a multidimensional objects, where the dimensions of the space are linked with some principal parameters of the iris and the pupil; then the algorithm executes the last analysis to recognize a person by using a genetic pipeline.

The result of this step is on the second row of the Fig.6. A tomography is performed too (see last row of Fig.6). This subunit is very innovative and is the core of the recognition.

**v)** After the analysis the data come back the DB, which is linked with a Dispatcher Information system. This unit automatically builds status report about people recognizing, people flows, doors and access of the buildings and so on. In particular, statistics, plots of data and events are produced and stored by this module. In the occurrence of a special events, like as an alert, this unit can automatically reach supervisors and the police with e-mail service and SMS (Short Message System). This module is put into practice by using the INTERNET TOOLKIT and the SPC tools by NI.

## Analysis and Results

The system has a trivial first safety level thanks to a code with five digit, that the user must type to start to recognizing procedure in supervised area. Starting from a single true colour image (see image 3) all the analysis is performed. Some analysis results are reported in Fig.6. It was test the accuracy of the system according to some crucial aspects. The system must recognize the same eye in different conditions:

1) the pupil can changes its dimension (i.e in a different light condition or because of the assumption of alcoholic drinks);
2) the sclera (white background of the eye) can appear more red because of congestion;
3) the curvature and other morphological parameters of the eye and iris can change in some pathologies, although these effects are foreseeable. They can be treated with genetic algorithms.

All These questions were considered in the pipeline. Then it is not so relevant the number of different pixels of the present image respect to the reference one in the DB, but it is really very important: i) where the pixels are in the image, if they are clustered, the morphology of the objects outside them, their colour, and their frequencies in some transformed spaces, their capability two changes respect to the time and respect to different condition of the data taking.

## Conclusion

An high challenge for safety is reach and an interesting solution is fixed. The present system get high level results in eye recognition. At the present no one has reach comparable results. The accuracy and the precision are different order up to the other present systems. IRIS System is scalable respect to the needed level of security and the budget. It is low cost system respect to the standard cost per year in public structure or enterprise.

## Acknowledgements

The author wish to thank I.Piacentini and the NI imaging group in Italy for relevant software suggestions and comments.

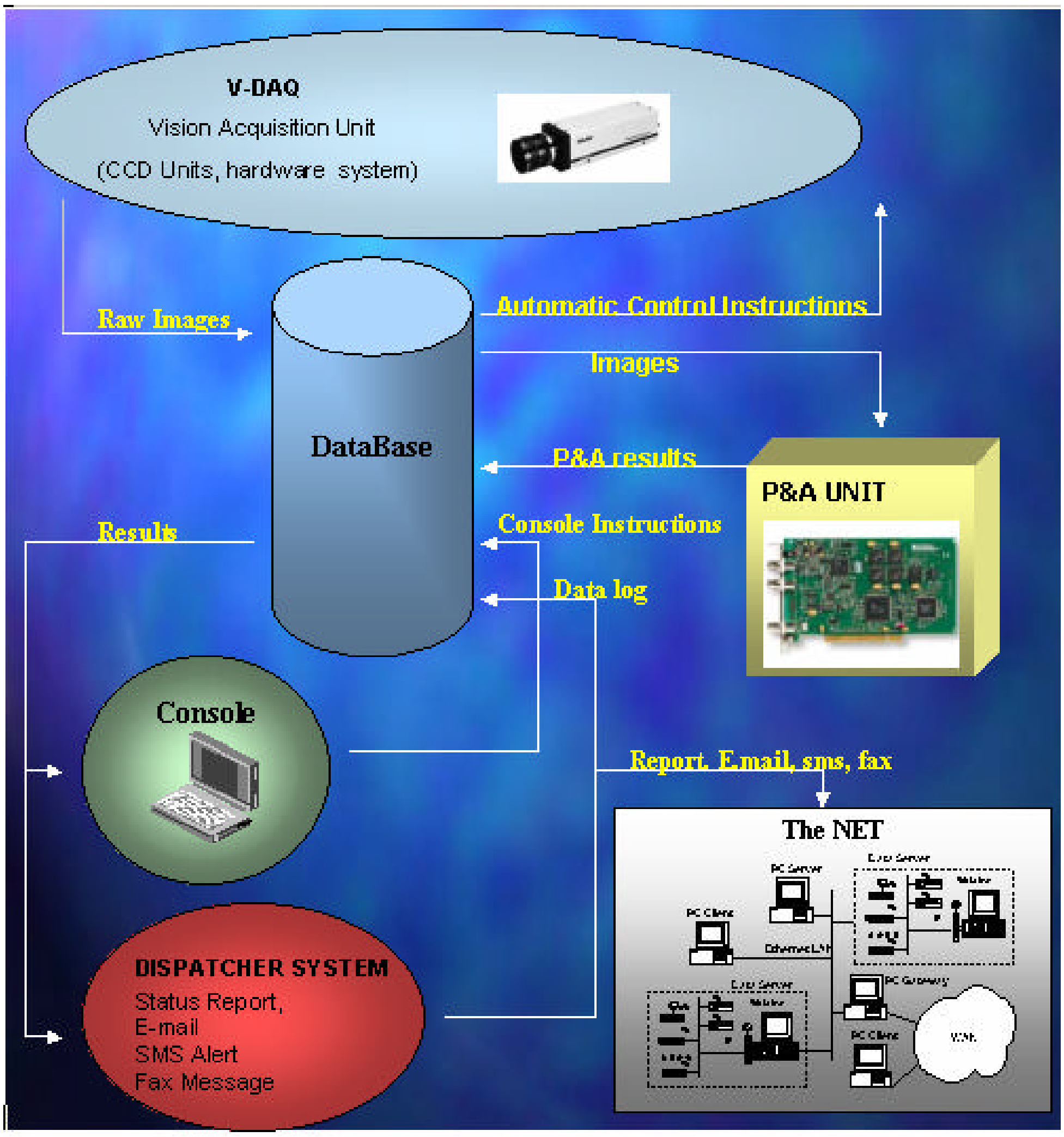

Fig.1: The system architecture.

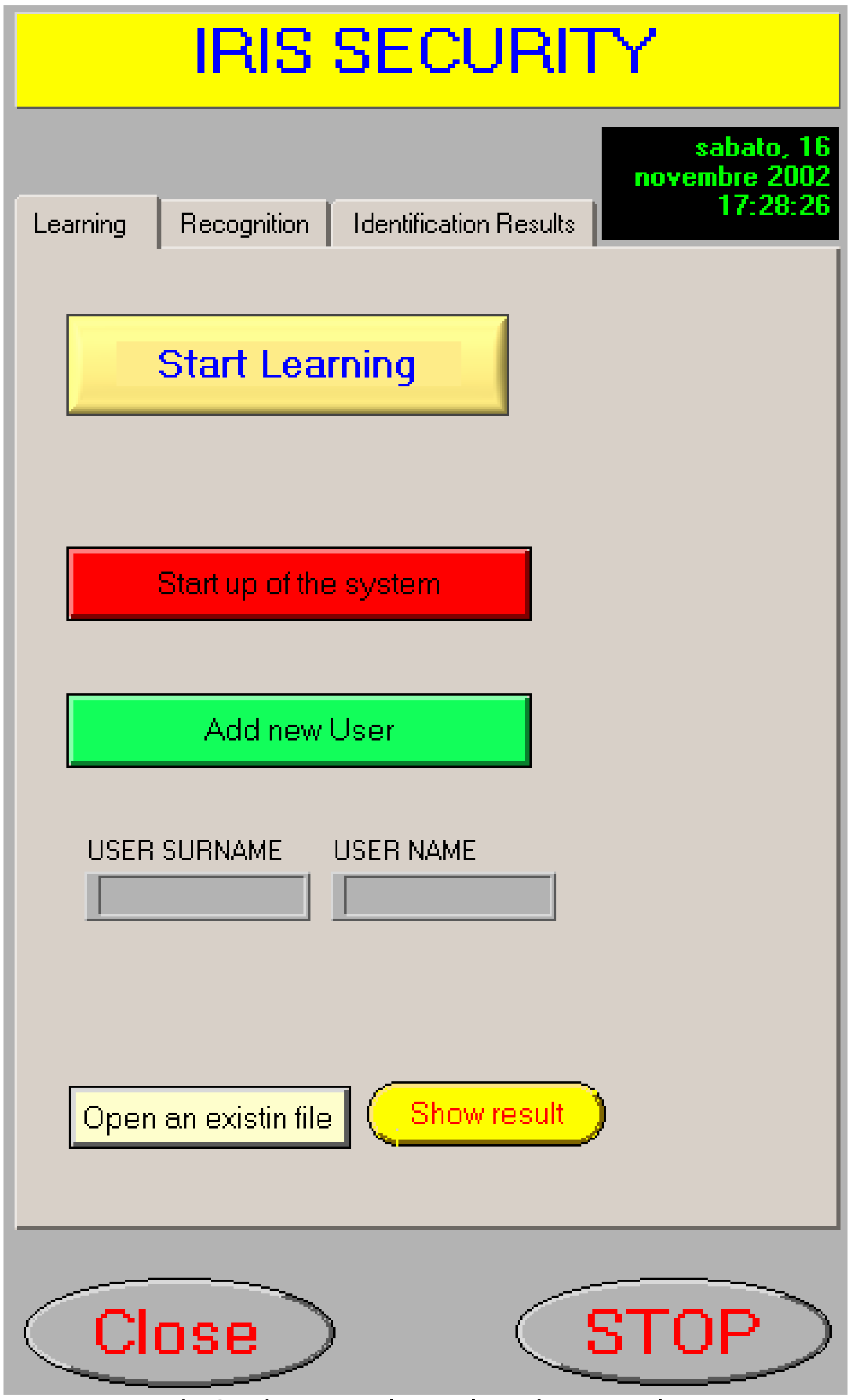

Fig.2: The control panel on the console.

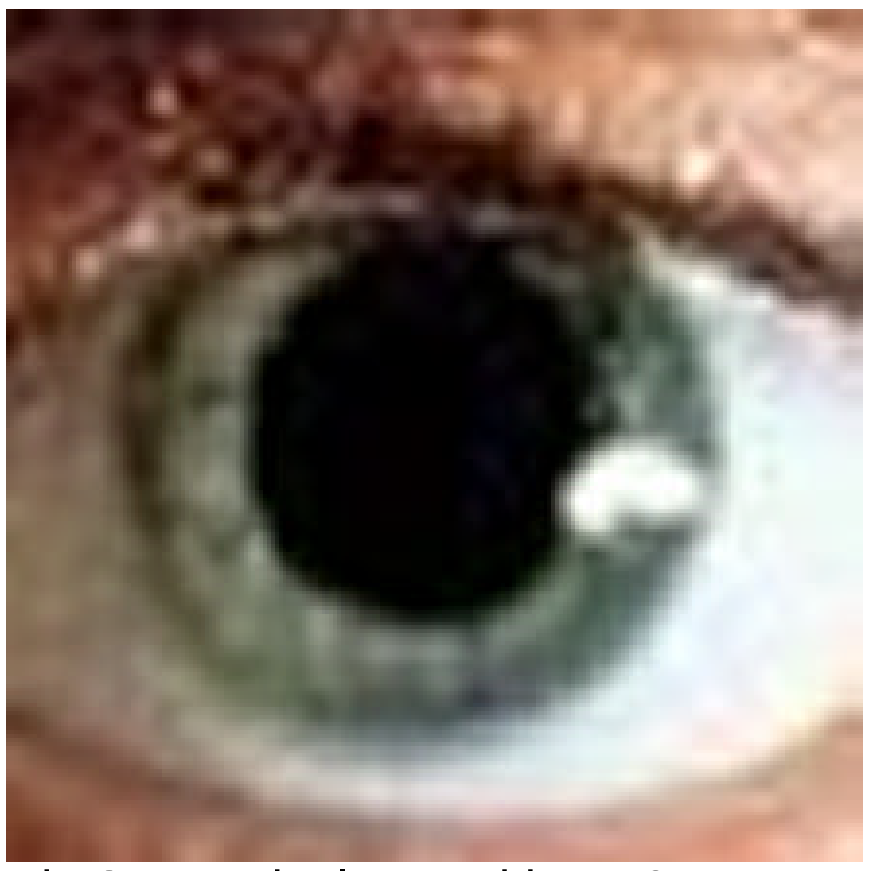
Fig.3: A typical eye with a 10X zoom.

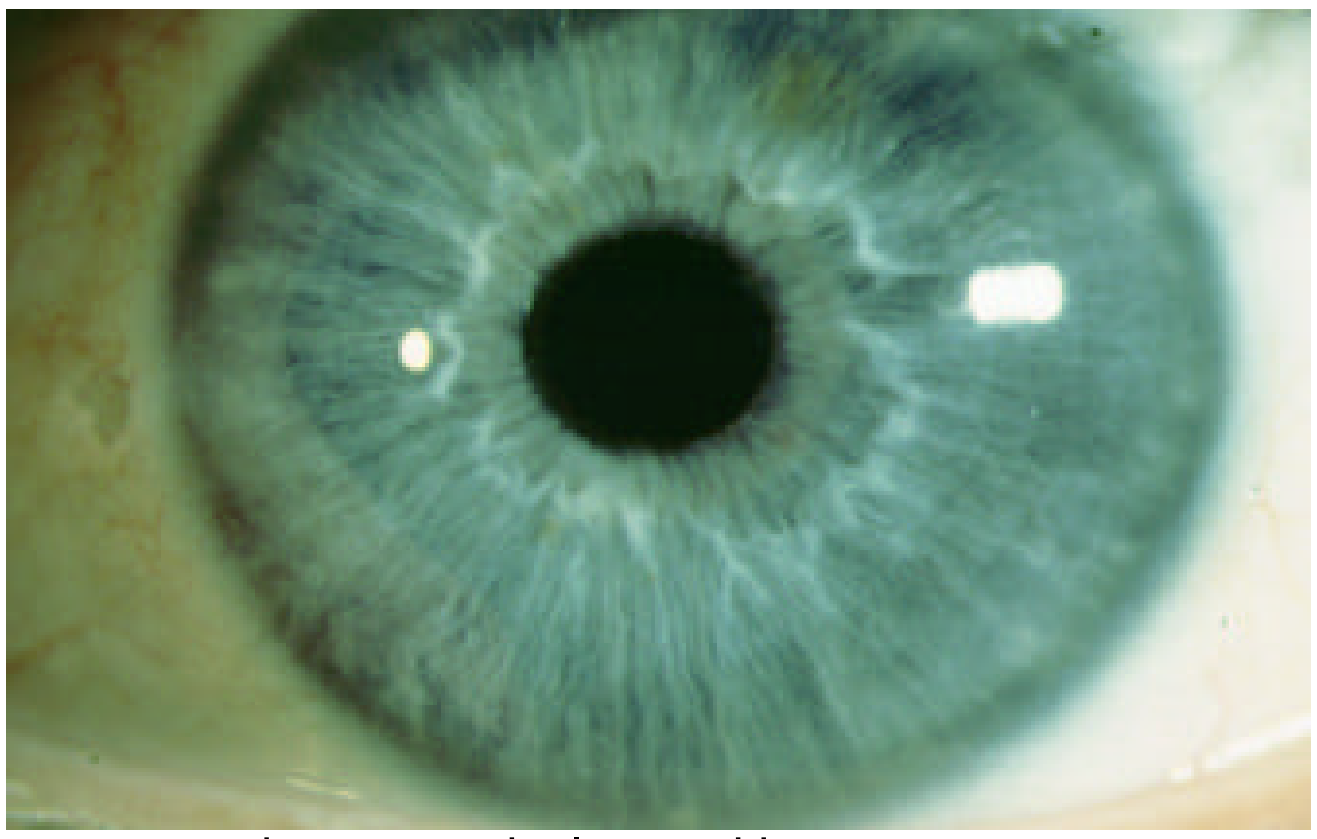
Fig. 4: A typical eye with a 15X zoom.

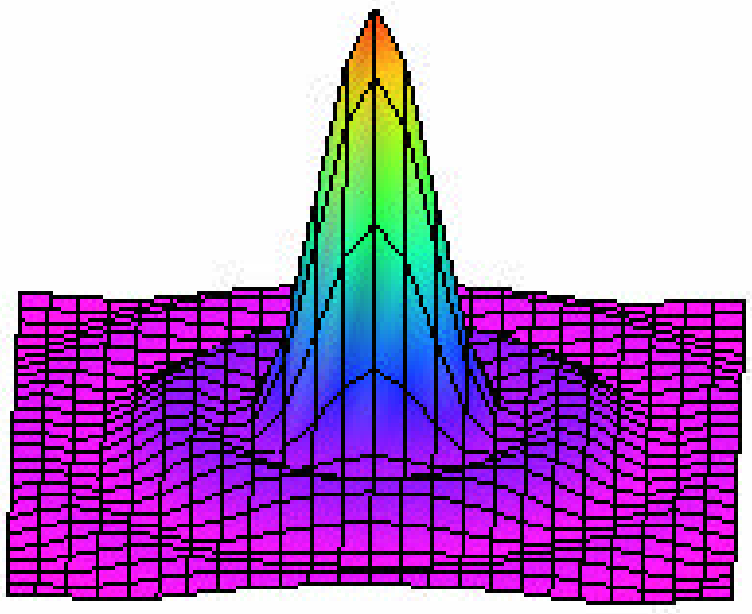
Fig.5: A 3-d iris in a transformed space.

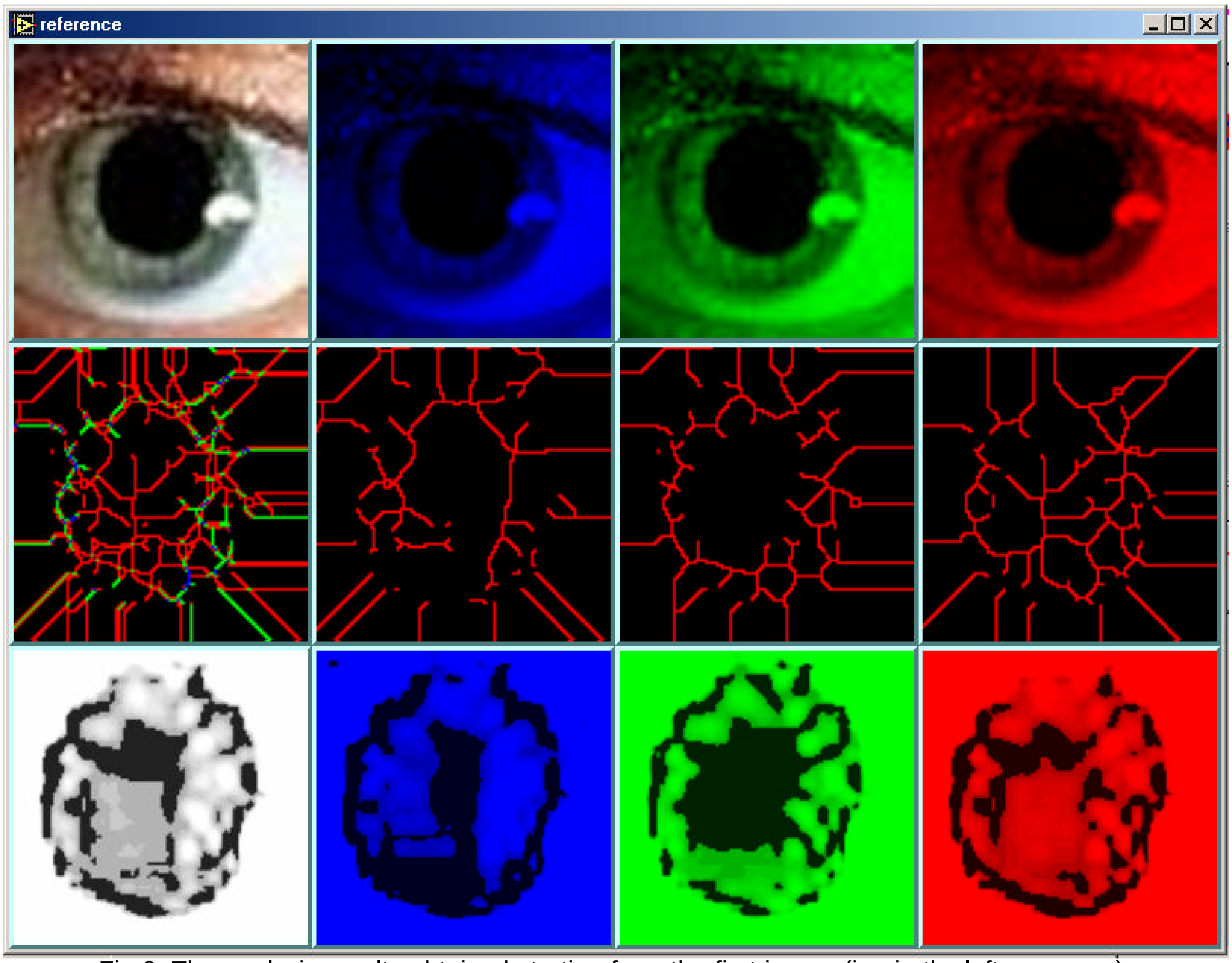

Fig.6: The analysis results obtained starting from the first image (i.e. in the left up corner).